\documentclass[letterpaper]{article} % DO NOT CHANGE THIS
\usepackage[preprint]{aaai2027}  % DO NOT CHANGE THIS
\usepackage[hyphens]{url}  % DO NOT CHANGE THIS
\usepackage{graphicx} % DO NOT CHANGE THIS
\usepackage{natbib}  % DO NOT CHANGE THIS AND DO NOT ADD ANY OPTIONS TO IT
\usepackage{caption} % DO NOT CHANGE THIS AND DO NOT ADD ANY OPTIONS TO IT
\usepackage{amsmath} 
\usepackage{amssymb}
 \usepackage{array} 
 \usepackage{booktabs}
 \usepackage{algorithm}
\usepackage{algorithmic}
\usepackage{amsfonts} 
\usepackage{multirow}
\newtheorem{lemma}{Lemma}

\newtheorem{theorem}{Theorem}
\title{Trace2Tower: Transition-Aware EigenTrace Induction of Multi-Level Skills for LLM Agents{}}

\author{
    Jiazheng Sun\equalcontrib\textsuperscript{\rm 1},
    Boyu Yang\equalcontrib\textsuperscript{\rm 1},
    Binhao Yuan\textsuperscript{\rm 1},
    Mingxuan Li\textsuperscript{\rm 1},
    Xin Peng\textsuperscript{\rm 1}
}

\affiliations{
    \textsuperscript{\rm 1}College of Computer Science and Artificial Intelligence, Fudan University\\
    Shanghai 200433, China
}

\begin{document}

\maketitle

\begin{abstract}
Large language model agents increasingly rely on execution traces to master complex interactive tasks. However, current paradigms are bottlenecked by shallow trajectory retrieval and flat skill summarization, fundamentally ignoring the temporal dependencies and outcome-conditioned topology of agent behavior. We introduce Trace2Tower, a transition-aware EigenTrace framework that distills raw trajectories into a robust skill hierarchy. Trace2Tower abstracts step-level interactions into canonical events, constructing a unified graph governed by semantic compatibility, transition dynamics, and outcome evidence. Through a novel contrastive spectral decomposition, it isolates stable, success-aligned behavioral modes while rigorously suppressing failure-prone shortcuts. These modes organically populate a dynamic skill tower of action templates, procedural routines, and overarching task strategies, continuously refined via verifier-guided feedback. On ALFWorld, Trace2Tower achieves 87.31\% success requiring only 10.35 steps and 0.26 invalid actions; on WebShop, it reaches 50.67\% exact success. Across both benchmarks, Trace2Tower significantly outperforms existing baselines in task mastery and context-efficient experience reuse.
\end{abstract}

\begin{links}
    \link{Code}{https://github.com/FudanSELab/Trace2Tower}
\end{links}


\section{Introduction}

\begin{figure*}[t]
    \centering
    \includegraphics[width=\textwidth]{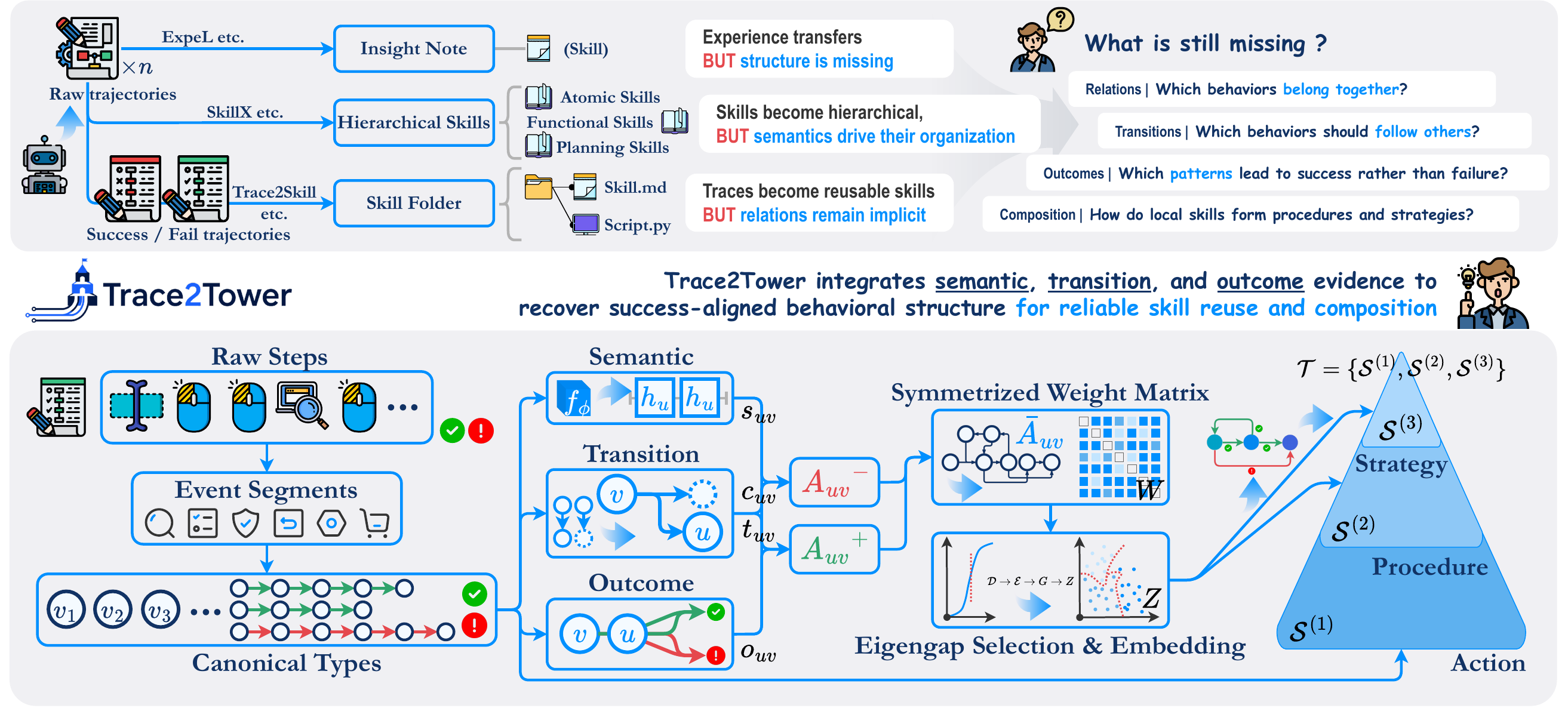}
    \caption{From experience reuse to behavioral structure induction. Existing methods transform trajectories into reusable insights or skills, but the underlying behavioral relations remain implicit. Trace2Tower recovers success-aligned behavioral structures through semantic, transition, and outcome-aware EigenTrace induction.}
    \label{fig:motivation}
\end{figure*}

LLM agents are increasingly expected to improve through continued interaction rather than address each task in isolation \cite{song2024eto,wang2026sage}. In interactive environments, agent behavior develops through a sequence of observations, actions, state changes, and feedback signals \cite{hu2025hiagent}. The value of an action therefore depends not only on its immediate effect, but also on how it shapes later decisions and the final outcome. Historical trajectories preserve this decision process and provide a rich source of behavioral knowledge beyond the pretrained model \cite{song2024agentbank}. They reveal effective action patterns, useful intermediate states, recovery procedures, and long-term strategies that support successful execution \cite{fu2024autoguide,agashe2024agent}. However, raw trajectories are difficult to reuse directly. They are often lengthy, redundant, and closely tied to specific tasks \cite{hu2025hiagent,yang2025cfgm}. They may also contain unnecessary exploration and unsuccessful behavioral branches that should not be reused indiscriminately \cite{song2024eto}. The central challenge is therefore to transform diverse execution traces into compact and reliable behavioral structures that retain decision patterns while removing task-specific noise and failure-related shortcuts.

Existing studies mainly address this challenge through experience reuse and skill construction. Experience-based methods preserve interactions as demonstrations, memories, guidelines, or retrieved experiences and use them to guide future execution \cite{fu2024autoguide,agashe2024agent,yang2025cfgm}. Skill-based methods further convert accumulated experience into persistent and reusable capabilities organized in skill libraries \cite{wang2026sage}. These directions have shown that historical interactions can improve task success, adaptation, and execution efficiency \cite{fu2024autoguide,wang2026sage}. Nevertheless, most existing representations remain centered on the content or organization of individual experiences, guidelines, plans, or skills \cite{fu2024autoguide,yang2025cfgm,wang2026sage}. They focus primarily on what information should be retained and reused, while behavioral relations among these units remain implicit. Events with similar descriptions may serve different functions during execution, whereas events with different surface forms may support the same procedure \cite{pysklo2026agent}. Such representations do not explicitly distinguish reliable execution transitions from failure-related shortcuts or explain how local behaviors compose into success-aligned procedures and task-level strategies \cite{yu2026does}.

These limitations have gradually shifted research from the reuse of complete trajectories toward the construction of more general and persistent skills. ExpeL extracts transferable insights from accumulated experience and combines them with relevant demonstrations to support future tasks \cite{zhao2024expel}. SkillX organizes reusable capabilities at multiple levels of abstraction for long-term agent learning \cite{wang2026skillx}. Trace2Skill further uses both successful and failed executions to extract trajectory-level lessons and consolidate them into reusable skill directories \cite{ni2026trace2skill}. This progression from raw experience to structured skill artifacts is reflected in Figure~\ref{fig:motivation}. Although these methods improve the organization and reuse of experience, the behavioral structure that supports skill formation remains largely implicit. They do not explicitly identify stable functional groups, reliable execution transitions, or recurring patterns that are associated with success rather than failure. They also provide limited support for modeling how local skills form reusable procedures and higher-level strategies. These limitations motivate a structure-aware method that jointly captures semantic relations, execution transitions, task outcomes, and hierarchical composition.

To address this problem, we propose Trace2Tower, a transition-aware EigenTrace framework that transforms raw LLM agent trajectories into structured multi-level skills. Trace2Tower abstracts executions into canonical events, constructs an outcome-aware transition graph, and applies contrastive spectral decomposition to discover success-aligned behavioral modes. These modes are organized into an action-, procedure-, and strategy-level skill tower with verifier-guided refinement. Experiments on ALFWorld and WebShop show that Trace2Tower achieves 87.31\% task success and 50.67\% exact success, while reducing ALFWorld execution steps to 10.35 and invalid actions to 0.26.
The main contributions of this paper are summarized as follows:

\begin{itemize}
    \item We propose Trace2Tower, a framework that transforms LLM agent trajectories into structured skills by explicitly capturing behavioral relations, execution dependencies, and outcome information beyond conventional skill extraction and experience accumulation approaches.

    \item We introduce transition-aware EigenTrace induction, which integrates semantic compatibility, transition dynamics, and success/failure evidence into a unified graph and employs contrastive spectral decomposition to uncover stable outcome-aligned behavioral structures.

    \item We develop a hierarchical skill construction and refinement mechanism that organizes induced behaviors into action-, procedure-, and strategy-level skills, enabling compact, transferable, and reliable experience reuse for LLM agents across long-horizon decision-making tasks.
\end{itemize}

%You can remove the copyright notice and ensure that your names aren't %shown by including \texttt{submission} option when loading the %\texttt{aaai2027} package:

%\begin{quote}\begin{scriptsize}\begin{verbatim}
%\documentclass[letterpaper]{article}
%\usepackage[submission]{aaai2027}
%\end{verbatim}\end{scriptsize}\end{quote}

%The remainder of this document are the original camera-
%ready instructions. Any contradiction of the above points
%ought to be ignored while preparing anonymous submis-
%sions.

\begin{figure*}[t]
    \centering
    \includegraphics[width=\textwidth]{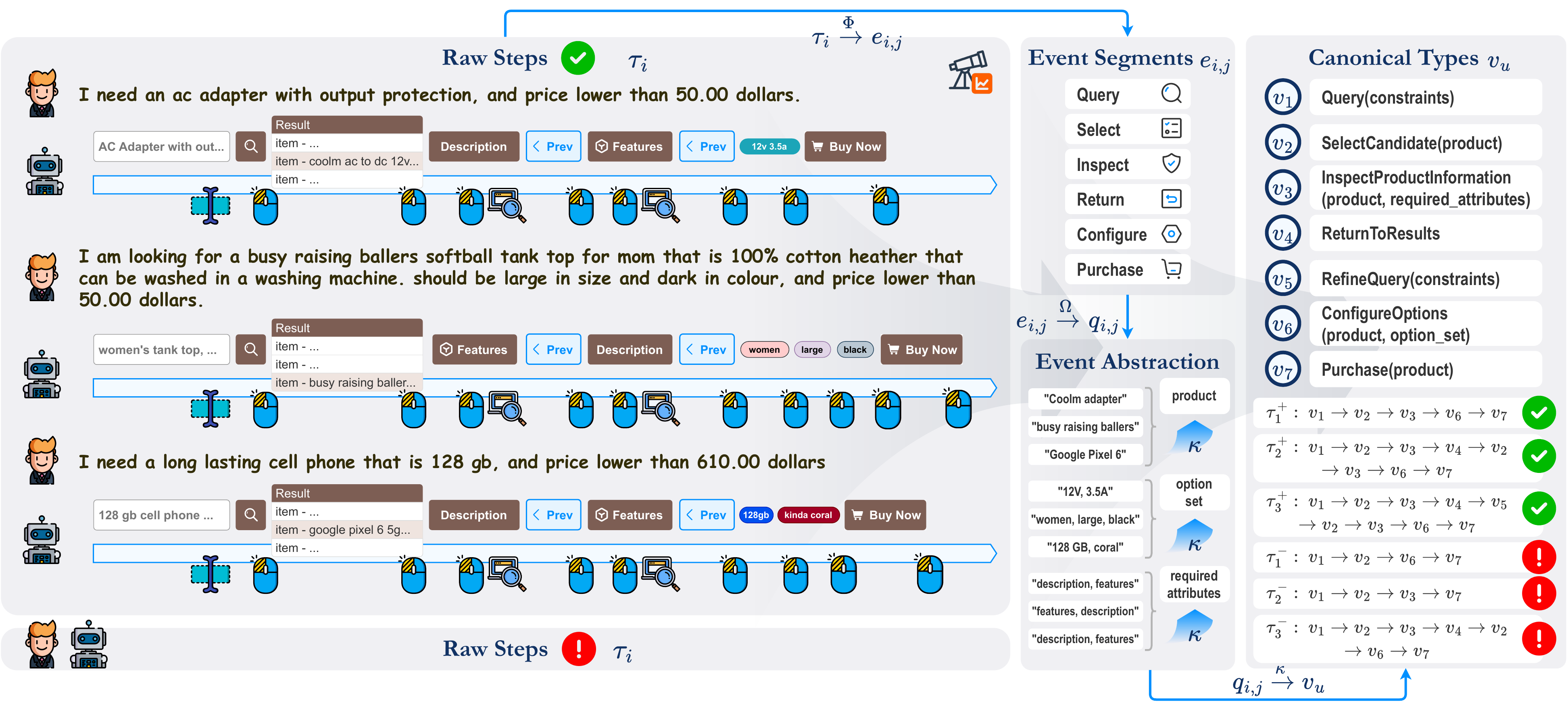}
    \caption{Event-level abstraction from raw agent trajectories. Trace2Tower converts successful and failed step-level executions into ordered canonical event sequences for subsequent behavioral structure induction.}
    \label{fig:event-abstraction}
\end{figure*}

\section{Related Work}

\subsubsection{Agent Experience Learning}
LLM agents reuse prior interactions through interfaces. ReAct exposes reasoning in the action loop; Reflexion converts feedback into persistent guidance; Voyager stores executable programs; ExpeL combines cross-task insights with retrieved episodes; and SEER retrieves step-level evidence from successful trajectories \cite{yao2022react,shinn2023reflexion,wang2023voyager,zhao2024expel,cui2025seer}. Memory studies further show that similarity-based experience following can propagate errors or replay misaligned experiences \cite{xiong2025memory}. These methods mainly reuse textual memories, demonstrations, trajectories, or artifacts selected by contextual relevance. Such representations do not explicitly preserve which transition enabled success, which branch failed, or which prerequisites preceded later actions. Trace2Tower instead induces these dependencies from trajectories.

\subsubsection{Skill Library Construction}
Recent work builds persistent repositories and structured skill graphs for lifelong learning. SkillFlow automates skill discovery; benchmarks assess acquisition and transfer over task streams \cite{zhang2026skillflow,li2026skillsbench,zheng2025lifelongagentbench}; SkillX refines hierarchical skills \cite{wang2026skillx}; and SAGE accumulates skills through sequential interaction \cite{wang2026sage}. Graph of Skills, SkillGraph, and SkillDAG further model inter-skill dependencies for retrieval and evolution \cite{li2026graphofskills,li2026skillgraph,bai2026skilldag}. These systems improve curation and selection, but structure is largely defined among existing skill artifacts. They do not explain how recurring execution events form procedures or why compositions align with success. Trace2Tower instead induces levels from semantic, temporal, and outcome relations.

\subsubsection{Structured Skill Induction}
Trace2Skill distills successful and failed trajectories into reusable skill directories \cite{ni2026trace2skill}. Recent methods also make skill relations explicit: GraSP compiles skills into dependency graphs for structured execution, while SkillGraph evolves directed relations from trajectories and feedback \cite{xia2026grasp,li2026skillgraph}. These approaches organize skill-level structure, whereas Trace2Tower addresses structure before reusable procedures exist. It builds an outcome-conditioned graph over canonical events and contrasts successful and failed transitions before spectral decomposition. Procedures and strategies therefore inherit observed execution order and success alignment rather than relations imposed over an existing skill collection.

\begin{figure*}[t]
    \centering
    \includegraphics[width=\textwidth]{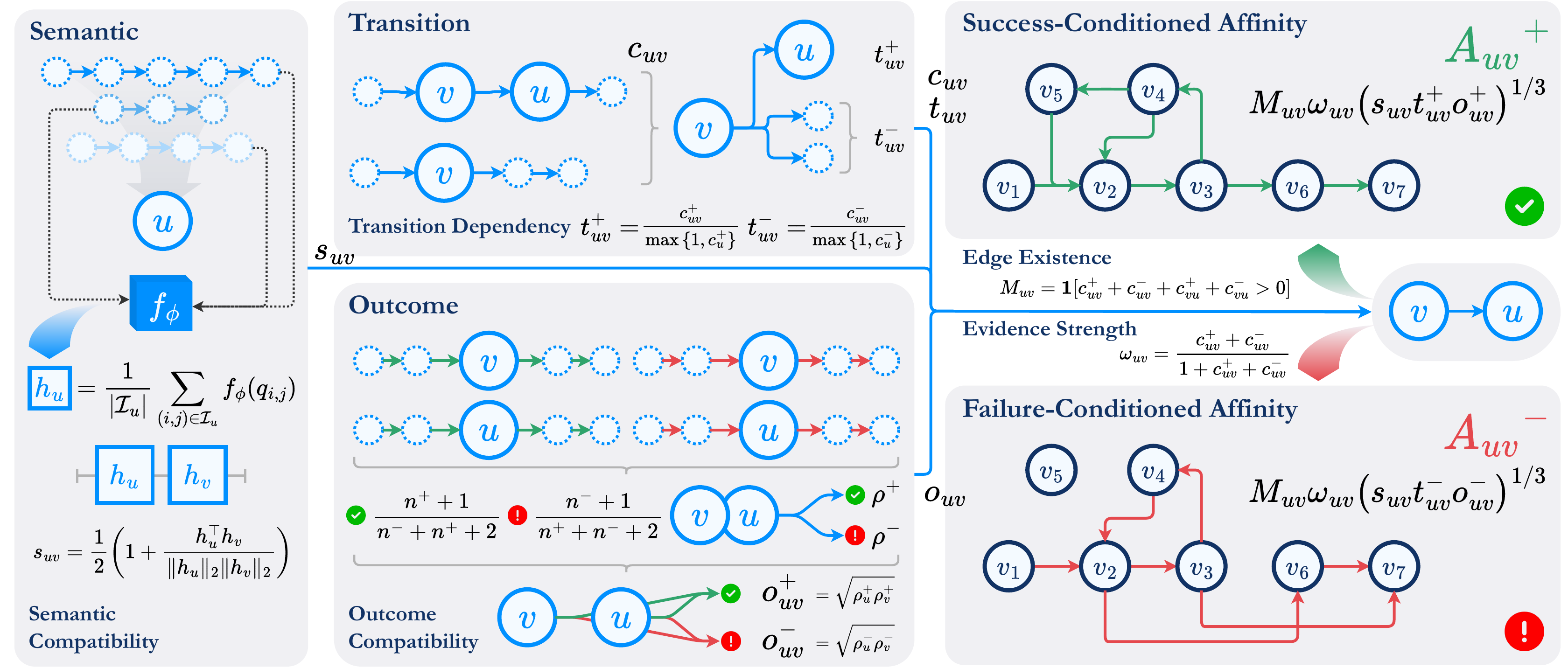}
    \caption{Construction of the transition-aware EigenTrace graph. The outcome-conditioned edge affinities $A_{uv}^{\pm}$ geometrically integrate semantic compatibility $s_{uv}$, transition dependency $t_{uv}^{\pm}$, and outcome compatibility $o_{uv}^{\pm}$, which are further gated by an edge mask $M_{uv}$ and scaled by evidence strength $\omega_{uv}$ to yield robust behavioral topologies.}
    \label{fig:eigentrace-graph}
\end{figure*}

\section{Problem Formulation}

Let $\mathcal{D}=\{(\tau_i,y_i)\}_{i=1}^{N}$ denote interaction histories, where $y_i\in\{0,1\}$ is task success and $\tau_i=\langle x_{i,t}\rangle_{t=1}^{T_i}$. Each $x_{i,t}=(g_i,o_{i,t},\mathcal{A}_{i,t},a_{i,t},f_{i,t})$ contains the task, observation, available actions, selected action, and feedback. Trace2Tower converts these fragmented steps into the ordered canonical events shown in Figure~\ref{fig:event-abstraction}:
\begin{equation}
\mathcal{E}_i
=
\Phi(\tau_i)
=
\langle e_{i,1},e_{i,2},\ldots,e_{i,M_i}\rangle,
\quad
\mathcal{E}
=
\{\mathcal{E}_i\}_{i=1}^{N}.
\label{eq:event-abstraction-overview}
\end{equation}
Each event represents a coherent execution stage while preserving sequence, state changes, and outcome evidence.
Given \(\mathcal{D}\), Trace2Tower learns a structure induction function \(\mathcal{F}:\mathcal{D}\rightarrow\mathcal{T}\), where \(\mathcal{T}=\{\mathcal{S}^{(1)},\mathcal{S}^{(2)},\mathcal{S}^{(3)}\}\) forms a three-level skill tower representing action-, procedure-, and strategy-level skills, respectively. The complete induction pipeline is formalized as:
\begin{equation}
\mathcal{D}
\rightarrow
\mathcal{E}
\rightarrow
G
\rightarrow
\mathcal{Z}
\rightarrow
\mathcal{T},
\label{eq:overall-pipeline}
\end{equation}
where \(G\) is the EigenTrace graph and \(\mathcal{Z}=\{Z^{(c)}\}_{c=1}^{C}\) contains spectral representations of its nontrivial components. The objective is a compact hierarchy whose relations are supported by semantic, transition, and outcome evidence.

\section{Method}

Trace2Tower distills accumulated LLM agent trajectories into a robust skill hierarchy via event abstraction, transition-aware EigenTrace construction, and parameter-free contrastive spectral decomposition. The extracted behavioral modes establish skills at the action, procedure, and strategy levels, which are dynamically refined through hierarchical retrieval and verifier feedback.

\subsubsection{Event-Level Trajectory Segmentation}

Trace2Tower groups consecutive steps that pursue a shared local objective into canonical events $\mathcal{E}_i=\Phi(\tau_i)$. Boundaries follow changes in subgoal, action family, manipulated entity, observed state, or feedback. Each event is summarized as $q_{i,j}=\Omega(g_i,e_{i,j})$, and task-specific entities are replaced by typed arguments before behaviorally equivalent summaries receive a common identity $\kappa(q_{i,j})=u$. The observed identities form graph nodes $\mathcal{V}=\{v_u\}_{u=1}^{m}$. With occurrence set $\mathcal{I}_u=\{(i,j)\mid\kappa(q_{i,j})=u\}$, their representation is $h_u=|\mathcal{I}_u|^{-1}\sum_{(i,j)\in\mathcal{I}_u}f_{\phi}(q_{i,j})$, reducing instance-level linguistic variation.

This separation preserves two kinds of invariance simultaneously. Typed arguments let the same operation recur across objects and tasks, while the original event order remains available for transition estimation. The graph can therefore share local behavior without flattening the temporal context that determines whether the behavior is executable.

\subsubsection{Transition-Aware EigenTrace Graph Construction}

The directed graph in Figure~\ref{fig:eigentrace-graph} combines three signals. Successful and failed transition counts $c_{uv}^{\pm}$ give $t_{uv}^{\pm}=c_{uv}^{\pm}/\max\{1,c_u^{\pm}\}$; mapped cosine similarity gives $s_{uv}\in[0,1]$; and trajectory-level occurrence counts give smoothed outcome tendencies $\rho_u^{\pm}=(n_u^{\pm}+1)/(n_u^{+}+n_u^{-}+2)$ and compatibility $o_{uv}^{\pm}=\sqrt{\rho_u^{\pm}\rho_v^{\pm}}$. The mask $M_{uv}=\mathbf{1}[c_{uv}^{+}+c_{uv}^{-}>0]$ retains observed transitions, and $\omega_{uv}=(c_{uv}^{+}+c_{uv}^{-})/(1+c_{uv}^{+}+c_{uv}^{-})$ discounts weak evidence. Their geometric integration requires no manual mixing weights:
\begin{equation}
A_{uv}^{\pm}
=
M_{uv}\omega_{uv}
\left(
s_{uv}t_{uv}^{\pm}o_{uv}^{\pm}
\right)^{\frac{1}{3}}.
\label{eq:outcome-conditioned-affinity}
\end{equation}
The matrices $A^{+}$ and $A^{-}$ therefore encode the same directed event relation under successful and failed outcomes.

The three signals play nonredundant roles. Semantics transfers evidence across linguistically varied events, transitions retain executable direction, and outcomes distinguish useful dependencies from failure-correlated shortcuts. Their geometric integration requires every retained edge to have joint support, preventing one high but isolated score from determining the behavioral topology.

\subsubsection{Contrastive EigenTrace Decomposition}

To suppress behavioral patterns prevalent in failures, Trace2Tower applies a parameter-free contrastive affinity transformation:
\begin{equation}
\bar{A}_{uv}
=
\begin{cases}
\dfrac{(A_{uv}^{+})^2}
{A_{uv}^{+}+A_{uv}^{-}},
&
A_{uv}^{+}+A_{uv}^{-}>0,
\\[6pt]
0,
&
\text{otherwise}.
\end{cases}
\label{eq:contrastive-affinity}
\end{equation}
This transformation increases with success affinity, decreases with failure affinity, and remains bounded by $A_{uv}^{+}$. Its robustness to estimated affinities is formalized below.

\begin{lemma}[Contrastive Stability]
For $x,y\geq0$, define $g(x,y)=x^2/(x+y)$ when $x+y>0$, and $g(0,0)=0$. Then $0\leq g(x,y)\leq x$, with $g$ nondecreasing in $x$ and nonincreasing in $y$. For estimated affinities $\widehat A^{+},\widehat A^{-}$ and their transformation $\widehat{\bar A}$,
\begin{equation}
\|
\widehat{\bar A}-\bar A
\|_F
\leq
\|
\widehat A^{+}-A^{+}
\|_F
+
\|
\widehat A^{-}-A^{-}
\|_F.
\label{eq:contrastive-matrix-bound}
\end{equation}
\end{lemma}
The lemma shows that failure suppression does not amplify total estimation error beyond the two input perturbations. Its proof is in Appendix~A.
Based on the directed contrastive matrix \(\bar A\), Trace2Tower constructs the symmetric affinity matrix \(W\), diagonal degree matrix \(D\), and normalized Laplacian \(L\):
\begin{equation}
\begin{gathered}
W
=
\frac{\bar{A}+\bar{A}^{\top}}{2},
D_{uu}
=
\sum_{v=1}^{m}W_{uv},
L
=
I-D^{-\frac{1}{2}}WD^{-\frac{1}{2}},
\end{gathered}
\label{eq:normalized-eigentrace-graph}
\end{equation}
where \(D_{uu}^{-1/2}=0\) for isolated nodes. Zero-degree nodes form singleton procedure groups, whereas positive-degree nodes are partitioned into disjoint connected components \(\mathcal{V}^{\circ}=\bigcup_{c=1}^{C}\mathcal{V}_c\). Spectral decomposition is executed independently per component; components smaller than three nodes default to single procedure groups.

For each component \(c\) with \(m_c=|\mathcal{V}_c|\geq3\), let \(L^{(c)}\) be its normalized Laplacian with sorted eigenvalues \(0=\lambda^{(c)}_1<\lambda^{(c)}_2\leq\cdots\leq\lambda^{(c)}_{m_c}\). Excluding the trivial degree-weighted direction \(q^{(c)}_1\), Trace2Tower dynamically determines the optimal procedure group count via the maximum eigengap:
\begin{equation}
r_c
=
\arg\max_{2\leq k\leq m_c-1}
\left(
\lambda^{(c)}_{k+1}
-
\lambda^{(c)}_{k}
\right).
\label{eq:eigengap-selection}
\end{equation}

The retained spectral basis and normalized EigenTrace representations are formulated as:
\begin{equation}
\begin{gathered}
Q^{(c)}
=
[q^{(c)}_2,q^{(c)}_3,\ldots,q^{(c)}_{r_c}]
\in\mathbb{R}^{m_c\times (r_c-1)},
\\
z_u
=
\frac{Q^{(c)}_{u,:}}
{\|Q^{(c)}_{u,:}\|_2},
\quad
u\in\mathcal{V}_c,
\qquad
Z^{(c)}
=
[z_u^{\top}]_{u\in\mathcal{V}_c}.
\end{gathered}
\label{eq:eigentrace-representation}
\end{equation}
Nodes with zero-norm representations form singleton groups, while $\mathcal{Z}=\{Z^{(c)}\}_{c=1}^{C}$ captures the remaining success-aligned topology. The retained eigenvectors solve the normalized spectral relaxation on each component, as characterized in Appendix~B. Their empirical stability depends on the separation selected by the eigengap.

\begin{theorem}[EigenTrace Stability]
Let $\|\widehat L^{(c)}-L^{(c)}\|_2\leq\varepsilon_c$ and let $\Delta_c=\min\{\lambda^{(c)}_2-\lambda^{(c)}_1,\lambda^{(c)}_{r_c+1}-\lambda^{(c)}_{r_c}\}$. If $\Delta_c>2\varepsilon_c$, then
\begin{equation}
\|
\sin\Theta
(\widehat Q^{(c)},Q^{(c)})
\|_F
\leq
\frac{2\sqrt{r_c-1}\varepsilon_c}{\Delta_c}.
\label{eq:subspace-bound}
\end{equation}
\end{theorem}
Thus, well-separated behavioral modes remain stable when graph statistics are estimated from finite traces; the exact spectral characterization and proofs are deferred to Appendix~C.

\begin{figure}[t]
    \centering
    \includegraphics[width=\columnwidth]{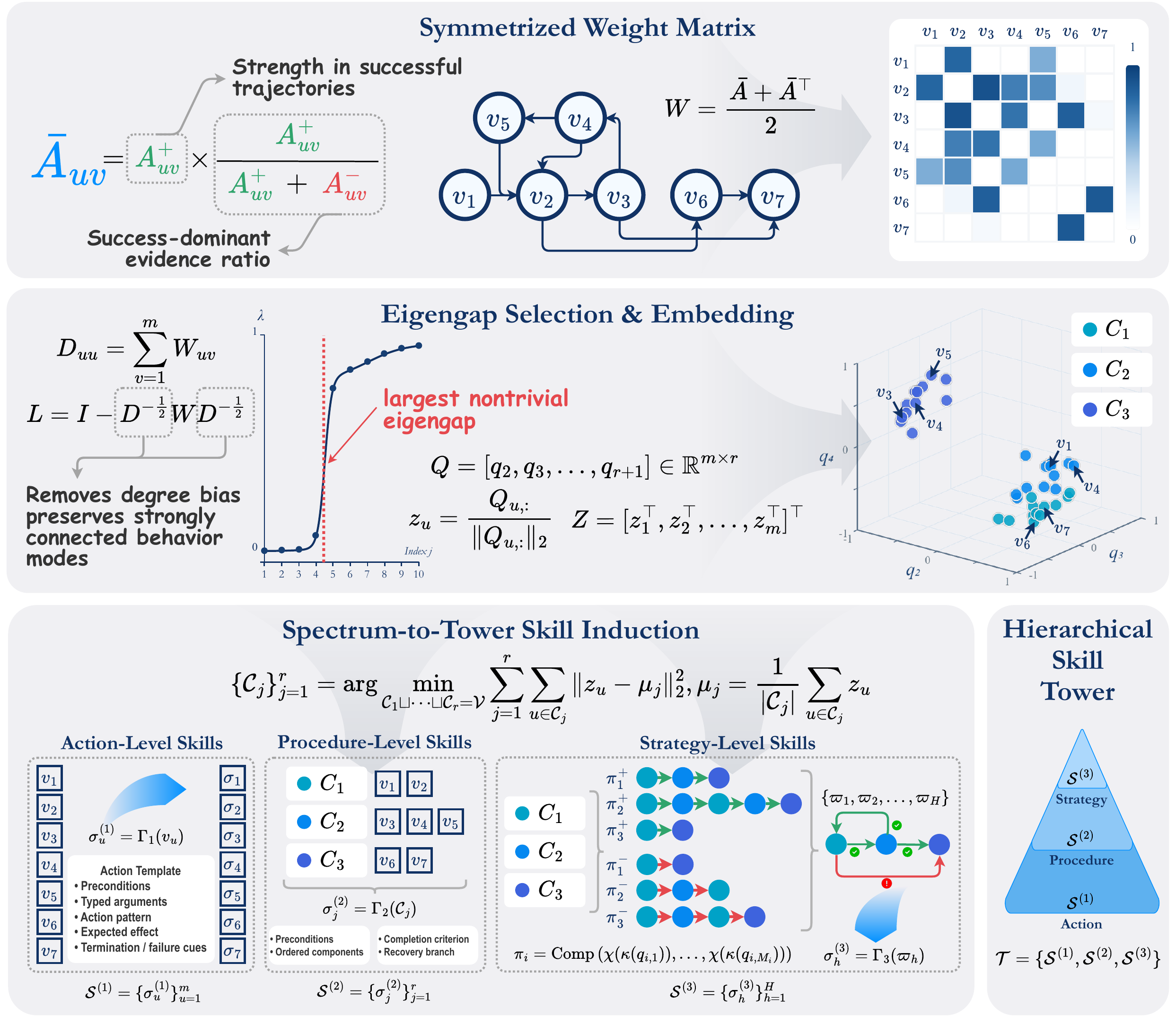}
    \caption{From success-aligned EigenTrace decomposition to hierarchical skill tower induction.}
    \label{fig:spectrum-to-tower}
\end{figure}

\begin{table*}[!t]
\centering
\small
\setlength{\tabcolsep}{2.2pt}
\renewcommand{\arraystretch}{1.15}

\begin{tabular*}{\textwidth}{@{\extracolsep{\fill}}
>{\raggedright\arraybackslash}p{3.15cm}
*{8}{c}
@{}}

\toprule

\multirow{2}{*}{\centering Method}
&
\multicolumn{5}{c}{ALFWorld}
&
\multicolumn{3}{c}{WebShop}
\\

\cmidrule(lr){2-6}
\cmidrule(lr){7-9}

&
\shortstack{Success\\\% $\uparrow$}
&
\shortstack{Steps\\$\downarrow$}
&
\shortstack{Invalid\\$\downarrow$}
&
\shortstack{Input\\Tokens $\downarrow$}
&
\shortstack{Skill Context\\Chars $\downarrow$}
&
\shortstack{Exact Success\\\% $\uparrow$}
&
\shortstack{Reward\\$\uparrow$}
&
\shortstack{Invalid\\$\downarrow$}
\\

\midrule

No-Skill
&
$46.27 \pm 5.83$
&
15.58
&
0.58
&
49,052
&
0
&
$40.67 \pm 0.58$
&
$0.6776 \pm 0.0044$
&
$0.68 \pm 0.03$
\\

Expert-Crafted
&
$71.39 \pm 4.11$
&
12.01
&
0.38
&
41,502
&
3,286
&
$48.00 \pm 0.00$
&
$0.7183 \pm 0.0000$
&
$0.07 \pm 0.02$
\\

\midrule

Trace2Skill +Combined
&
$61.44 \pm 2.15$
&
13.91
&
0.28
&
67,186
&
9,508
&
$40.67 \pm 0.58$
&
$0.5580 \pm 0.0084$
&
$2.39 \pm 0.07$
\\

Trace2Skill +Error
&
$61.19 \pm 1.29$
&
14.19
&
0.37
&
73,447
&
11,447
&
$44.00 \pm 0.00$
&
$0.5892 \pm 0.0055$
&
$1.81 \pm 0.02$
\\

SkillX 
&
$78.61 \pm 3.53$
&
12.06
&
0.40
&
64,229
&
10,039
&
$\underline{49.33 \pm 2.08}$
&
$0.6922 \pm 0.0102$
&
$\underline{0.32 \pm 0.04}$
\\

ExpeL
&
$81.59 \pm 0.86$
&
11.50
&
0.31
&
52,456
&
7,469
&
$48.00 \pm 0.00$
&
$\underline{0.6987 \pm 0.0101}$
&
$0.41 \pm 0.05$
\\

\midrule

Trace2Tower High-only
&
$\underline{84.83 \pm 0.86}$
&
\underline{10.52}
&
\underline{0.27}
&
\textbf{37,649}
&
\textbf{1,966}
&
$48.33 \pm 0.58$
&
$0.6935 \pm 0.0116$
&
$0.40 \pm 0.06$
\\

Trace2Tower Full
&
$\mathbf{87.31 \pm 0.75}$
&
\textbf{10.35}
&
\textbf{0.26}
&
\underline{44,651}
&
\underline{3,623}
&
$\mathbf{50.67 \pm 2.89}$
&
$\mathbf{0.7115 \pm 0.0183}$
&
$\mathbf{0.27 \pm 0.05}$
\\

\bottomrule

\end{tabular*}
\caption{Overall performance and inference efficiency on ALFWorld and WebShop across three independent runs. Bold and underline denote the best and second-best automatic methods.}
\label{tab:main_results}
\end{table*}

\subsubsection{Hierarchical Skill Induction and Deployment}

Figure~\ref{fig:spectrum-to-tower} illustrates the skill tower $\mathcal{T}=\{\mathcal{S}^{(1)},\mathcal{S}^{(2)},\mathcal{S}^{(3)}\}$. Action skills map from canonical events. Procedure skills cluster normalized EigenTrace representations using the eigengap-selected count; zero-degree nodes and components smaller than three remain isolated groups. For strategy induction, trajectories are compressed into repetition-free procedure sequences, the contrastive graph is rebuilt over procedure groups, and strongly connected components are collapsed. Distinct maximal paths supported by observed successful contrastive edges become strategy skills. Exact clustering and path construction are provided in Appendix~D. During deployment, encoded context $\xi_t=f_{\psi}(g,o_t,a_{1:t-1})$ is matched to each skill's supporting-event representation $\bar h_{\sigma}$:
\begin{equation}
\zeta_t(\sigma)
=
\begin{cases}
\dfrac{1}{2} + \dfrac{\xi_t^{\top}\bar h_{\sigma}}{2\|\xi_t\|_2\|\bar h_{\sigma}\|_2},
&
\|\xi_t\|_2\|\bar h_{\sigma}\|_2>0,
\\[6pt]
0,
&
\text{otherwise}.
\end{cases}
\label{eq:skill-relevance}
\end{equation}
With successful ($n_{\sigma}^{+}$) and total ($n_{\sigma}$) invocation counts, empirical reliability is smoothed as $\varrho_{\sigma}=(n_{\sigma}^{+}+1)/(n_{\sigma}+2)$. Given a prompt-token cost $\ell_{\sigma}$, budget $B_t$, and dependency-closed skill universe $\mathcal{H}_t$, the optimal hierarchical skill set is retrieved via:
\begin{equation}
\mathcal{R}_t=\operatorname*{arg\,max}_{\substack{\mathcal{R}\in\mathcal{H}_t\\\sum_{\sigma\in\mathcal{R}}\ell_{\sigma}\leq B_t}}\sum_{\sigma\in\mathcal{R}}\log\left(1+\zeta_t(\sigma)\varrho_{\sigma}\right).
\label{eq:hierarchical-retrieval}
\end{equation}

Post-episode feedback updates utility from empirical reliability, reward improvement, step savings, and invocation cost. Structural evidence then governs splitting, merging, promotion, and down-weighting. Appendix~E specifies these operations, and Appendix~F gives the complete time and memory analysis; the dominant spectral computation operates component-wise on the sparse behavioral graph.

\section{Experiments}

\subsection{Experimental Setup}

ALFWorld \cite{2020alfworld} evaluates household manipulation tasks featuring explicit prerequisite dependencies and state transitions. We evaluate on all 134 solvable \texttt{valid\_unseen} tasks across six families, inducing skills from 1,240 No-Skill trajectories collected over 310 training tasks. WebShop \cite{yao2022webshop} assesses web-based search and purchasing behavior over a product catalog. Skills for WebShop are induced from 400 trajectories across 100 training tasks, with all methods benchmarked on a fixed 100-task test manifest. Both environments enforce an interaction horizon of 20 steps. For ALFWorld, we report success rate, execution steps, invalid action, input tokens, and skill-context character length; for WebShop, we evaluate exact success rate, average reward, and invalid actions.

Unless stated otherwise, GPT-5.4 serves as the Skill Author and plan rewriter, while DeepSeek-V4-Flash acts as the Skill User. We evaluate two deployment policies over the same Tower: the High-only policy rewrites three retrieved strategy paths into a compact plan, whereas the Full policy additionally injects up to eight step-aligned procedure skills. We compare against No-Skill, Expert-Crafted, ExpeL \cite{zhao2024expel}, SkillX \cite{wang2026skillx}, and both the +Combined and +Error variants of Trace2Skill under identical trajectory pools, Skill User, test tasks, and interaction budgets. Table~\ref{tab:main_results} reports the mean and standard deviation across three independent runs; detailed configurations are provided in Appendix~G.

\subsection{Overall Performance}

Table~\ref{tab:main_results} shows that both Tower policies consistently outperform baselines across benchmarks. On ALFWorld, High-only reaches 84.83\% success with minimal token and context overhead, while Full achieves a peak success of 87.31\%, beating ExpeL by 5.72 points and SkillX by 8.70 points with only 10.35 steps and 0.26 invalid actions per episode. The 2.48-point gap between them highlights the value of state-aligned procedural skills beyond a global strategy plan. On WebShop, Full achieves the top exact success rate of 50.67\%, outperforming SkillX by 1.34 points and cutting invalid actions from 0.68 in No-Skill to 0.27. High-only reaches 48.33\% exact success, showing that procedural skills add a 2.34-point boost when interaction constraints arise, while its 0.7115 reward leads automated methods and approaches the 0.7183 Expert-Crafted benchmark.

\begin{figure}[t]
\centering
\includegraphics[width=\columnwidth]{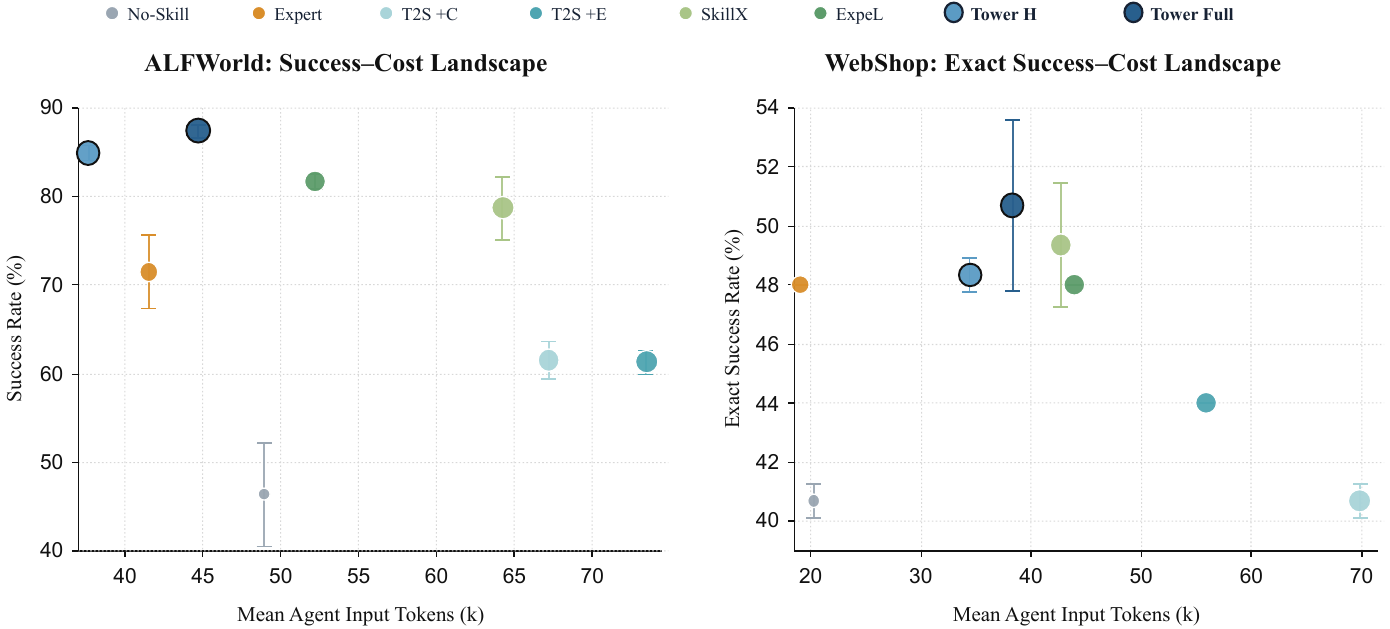}
\caption{Success--cost trade-offs on ALFWorld and WebShop over three runs.}
\label{fig:success_cost}
\end{figure}

Figure~\ref{fig:success_cost} illustrates that High-only provides cost efficiency while Full maximizes performance over the same Tower. Rather than a full method versus an ablation, they offer complementary deployment choices: High-only prioritizes concise long-horizon guidance, whereas Full uses context budget for state-conditioned execution. Efficiency metrics confirm this design. Compared with SkillX, Full cuts ALFWorld input tokens by 30.5\% and injected context by 63.9\%, while High-only further reduces these costs by 15.7\% and 45.7\% relative to Full. Thus, performance gains stem from structured skill organization rather than context window expansion.

\begin{figure}[t]
\centering
\includegraphics[width=\columnwidth]{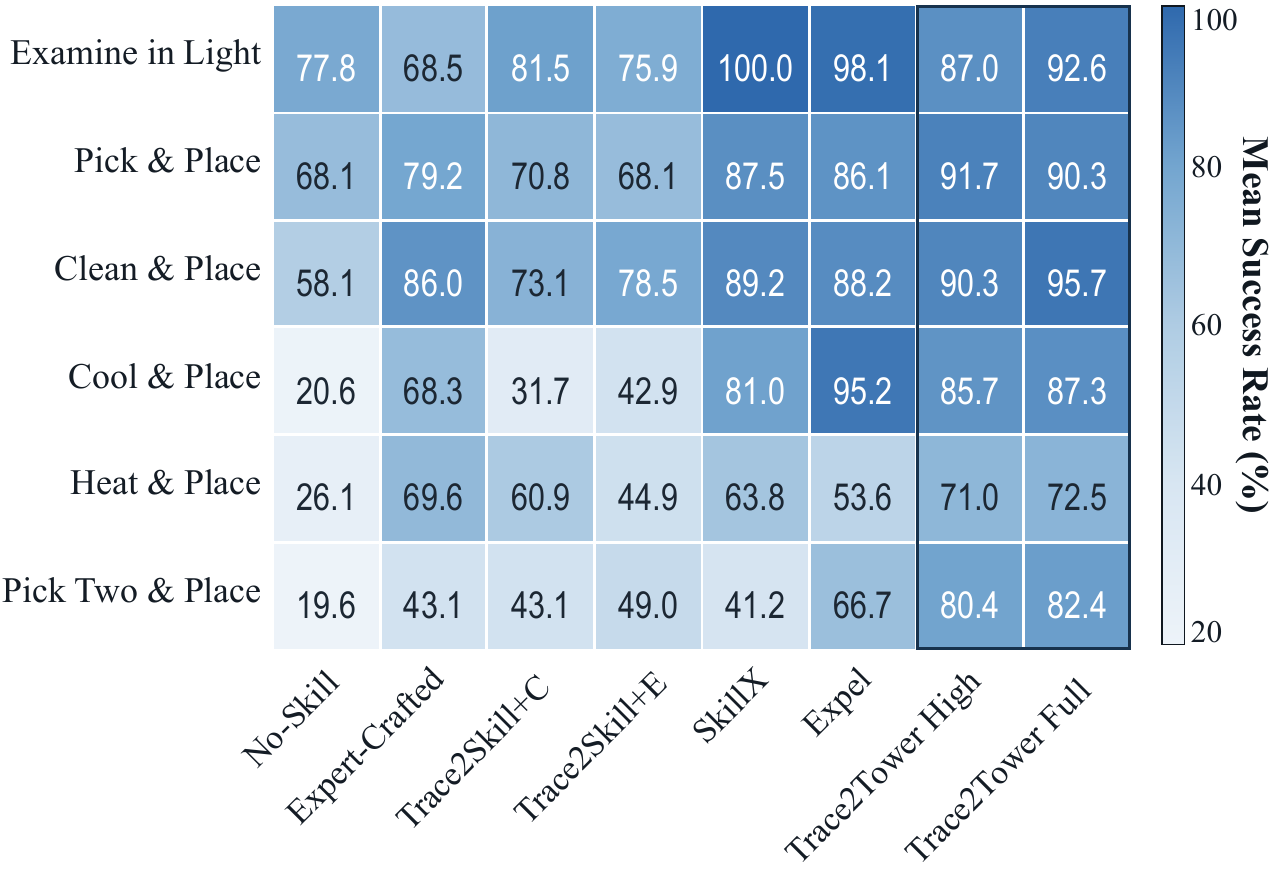}
\caption{ALFWorld success rates (\%) by task family and method. Outlined columns denote the two Trace2Tower deployment policies.}
\label{fig:task_family}
\end{figure}

\begin{figure}[t]
\centering
\includegraphics[width=\columnwidth]{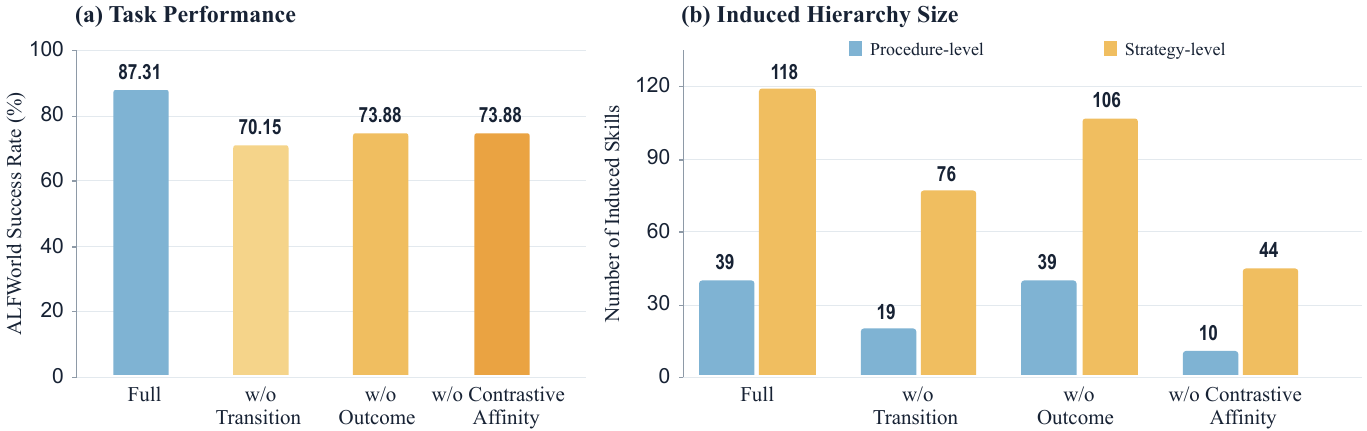}
\caption{Structural ablation on ALFWorld: (a) success rate and (b) induced procedure and strategy skills.}
\label{fig:graph_ablation}
\end{figure}

\begin{table}[t]
\centering
\small
\setlength{\tabcolsep}{4pt}
\renewcommand{\arraystretch}{1.0}
\begin{tabular}{lcc}
\toprule
\multirow{2}{*}{Experience}
& \multicolumn{2}{c}{Skill User} \\
\cmidrule(lr){2-3}
& \shortstack{DeepSeek-V4\\Flash}
& \shortstack{DeepSeek-V4\\Pro} \\
\midrule
No-Skill & 52.99 & 66.42 \\
\midrule
GPT-5.4 Tower & \textbf{88.06} & \textbf{85.82} \\
\shortstack[l]{DeepSeek-V4\\Flash Tower} & 65.67 & 79.10 \\
\bottomrule
\end{tabular}
\caption{ALFWorld success (\%) by experience condition and Skill User.}
\label{tab:author_user}
\end{table}

\begin{figure}[t]
\centering
\includegraphics[width=\columnwidth]{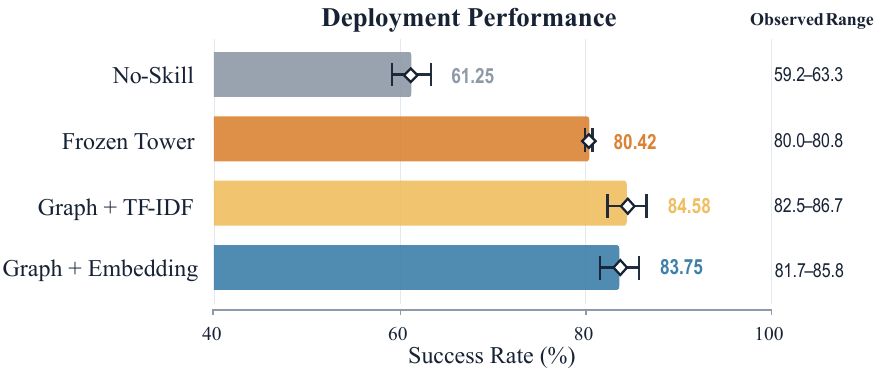}
\caption{Deployment success and observed range over two held-out ALFWorld sets.}
\label{fig:deployment_adaptation}
\end{figure}

\begin{figure}[t]
\centering
\includegraphics[width=\columnwidth]{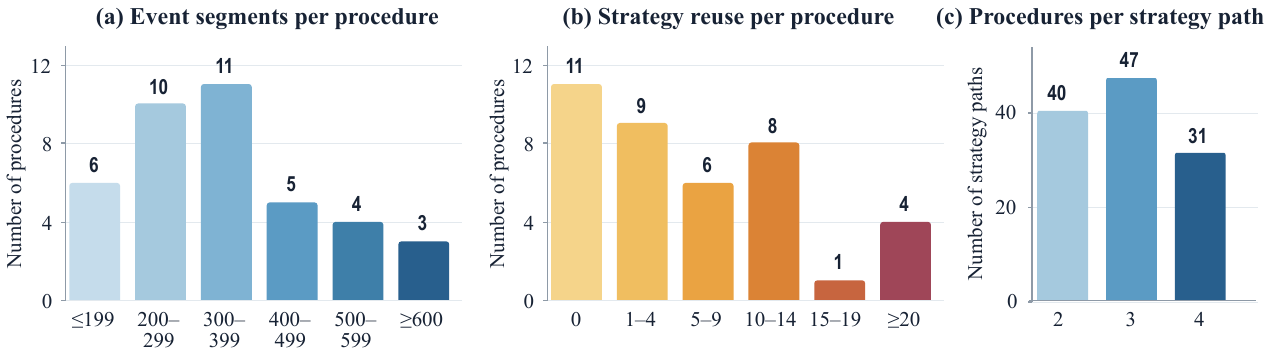}
\caption{Hierarchy organization: (a) supporting event segments per procedure, (b) strategy-path reuse per procedure, and (c) procedures per strategy path.}
\label{fig:hierarchy_mechanism}
\end{figure}

\subsection{Generalization and Structural Ablations}

Figure~\ref{fig:task_family} demonstrates that both Tower deployment policies maintain consistently high performance across all six task families on ALFWorld. Full achieves the top success rate on Clean-and-Place, Heat-and-Place, and Pick-Two-and-Place, whereas High-only leads on Pick-and-Place. In contrast, competing baselines exhibit pronounced performance degradation on specific task families despite strong overall averages. This stability indicates that Trace2Tower generalizes effectively across changing prerequisite chains, state transformations, and object multiplicity rather than relying on family-specific skills.

To isolate the contribution of each graph component, we perform structural ablations on ALFWorld, where high task diversity and explicit environment feedback facilitate fine-grained diagnosis. As shown in Figure~\ref{fig:graph_ablation}, removing transition dynamics causes the most severe performance drop, reducing success from 87.31\% to 70.15\% and shrinking the induced hierarchy from 39 procedures and 118 strategies down to 19 procedures and 76 strategies. Disabling outcome evidence or contrastive affinity lowers success to 73.88\%, with the latter collapsing the hierarchy to 10 procedures and 44 strategies. These results confirm that transitions encode necessary execution order, while outcome contrast prevents yet success-ambiguous relations from dominating the behavioral graph.

\subsection{Cross-Model Transfer and Feedback Refinement}

We evaluate cross-model transfer by decoupling the Skill Author from the Skill User. Table~\ref{tab:author_user} compares each user model under No-Skill control against Towers generated by two distinct author models. The Flash-authored Tower improves performance for both user models by 12.69 percentage points. For the Flash user, a GPT-5.4-authored Tower further raises success from 65.67\% to 88.06\%, achieving 34 paired wins against 4 losses with a p-value of $6.04\times10^{-7}$. For the Pro user, the margin between 79.10\% and 85.82\% yields a p-value of .150. These findings show that graph structure transfers robustly across model architectures, while author quality and user capability determine the realized gain.

Trace2Tower optimizes deployment from verifier feedback without reconstructing trajectories. Four graph edit operations split heterogeneous Mid evidence, merge compatible procedures, promote supported Mid paths, and downweight harmful High paths. Relation expansion connects candidate pairs, while Pareto selection balances semantic, child, and feedback relevance. As evaluated in Figure~\ref{fig:deployment_adaptation} across two held-out ALFWorld sets, updating the frozen Tower raises success from 80.42\% to 84.58\% with TF-IDF and 83.75\% with embeddings, confirming that feedback effectively revises future retrieval without re-inducing the Tower.

\subsection{Hierarchy Organization and Interface Stability}

Figure~\ref{fig:hierarchy_mechanism} demonstrates the functional separation across hierarchy levels. Each of the 39 induced procedures aggregates 352 event segments on average, while strategy paths compose only two to four procedures. Notably, 4 procedures recur in over 20 strategy paths whereas 11 appear in none. High-only thus concentrates on 28 recurrent procedures, whereas Full additionally retrieves specialized Mid-level procedures outside global paths, driving their quality-cost trade-off. Overall, Trace2Tower compresses 13,724 segments into 3,764 quotient nodes to induce 39 procedures and 118 strategies, consuming 16.85\% fewer GPT tokens during construction compared to SkillX.

We also evaluate stability across four experience interfaces, spanning fixed procedures in Trace2Skill, insights with episodes in ExpeL, queried multi-level skills in SkillX, and relational High and Mid composition in Trace2Tower. Interface stability closely tracks environment structure. In ALFWorld, upfront task family and state disclosures anchor the execution plan, keeping standard deviations for High-only and Full below one percentage point at 0.86 and 0.75 respectively. Conversely, WebShop reveals constraints dynamically through page transitions, increasing sensitivity to site traversal. Consequently, Full and SkillX yield higher standard deviations of 2.89 and 2.08, whereas High-only remains stable at 0.58, confirming that variance reflects dynamic interface sensitivity rather than poor skill transfer.

\section{Conclusion}

Trace2Tower induces a multi-level hierarchy from semantic, transition, and outcome evidence. It achieves 87.31\% ALFWorld success and 50.67\% WebShop exact success with efficient execution. Ablations and cross-model transfer validate the induced behavioral structure. Feedback-based graph edits further improve a frozen Tower without re-induction, demonstrating that the hierarchy remains editable after deployment rather than becoming a static skill library tied to its original trajectory pool. Together, these results support compact and adaptable skills across distinct execution environments.

\bibliography{aaai2027}

\clearpage

\appendix

\section{Appendix}
This supplementary material provides the complete mathematical proofs, the exact structure-induction procedure, the reported deployment policy, essential implementation settings, and additional analyses.

% ============================================================
% Appendix A
% ============================================================

\subsection{Appendix A: Proof of Contrastive Stability}
\label{app:contrastive_stability}

This appendix proves the Contrastive Stability lemma for the outcome-conditioned affinities used by Trace2Tower. For a directed canonical-event pair $(u,v)$, let $x=A_{uv}^{+}$ and $y=A_{uv}^{-}$ denote the success- and failure-conditioned affinities in Equation~\eqref{eq:outcome-conditioned-affinity}. Define
\begin{equation}
g(x,y)
=
\begin{cases}
\dfrac{x^2}{x+y}, & x+y>0,\\[4pt]
0, & x=y=0,
\end{cases}
\qquad x,y\geq0,
\label{eq:app-contrastive-scalar}
\end{equation}
so that $\bar A_{uv}=g(A_{uv}^{+},A_{uv}^{-})$.
For $x+y>0$, we have $g(x,y)=x\,x/(x+y)$. Since $0\leq x/(x+y)\leq1$, it follows that $0\leq g(x,y)\leq x$. Moreover, $\partial g/\partial x=x(x+2y)/(x+y)^2\in[0,1]$ and $\partial g/\partial y=-x^2/(x+y)^2\in[-1,0]$. Therefore, the retained contrastive edge is nondecreasing in success affinity and nonincreasing in failure affinity. The definition at $(0,0)$ is continuous because $0\leq g(x,y)\leq x\rightarrow0$.
Let $(x,y)$ and $(x',y')$ be two nonnegative affinity pairs. Applying the mean-value theorem along the two coordinates gives
\begin{equation}
\begin{aligned}
|g(x',y')-g(x,y)|
&\leq
|g(x',y')-g(x,y')|
\\
&\quad+
|g(x,y')-g(x,y)|
\\
&\leq
|x'-x|+|y'-y|.
\end{aligned}
\label{eq:app-entrywise-lipschitz}
\end{equation}

The same inequality holds on the boundary by continuity. Substituting
$x=A_{uv}^{+}$,
$y=A_{uv}^{-}$,
$x'=\widehat A_{uv}^{+}$, and
$y'=\widehat A_{uv}^{-}$ into Equation~\eqref{eq:app-entrywise-lipschitz} yields the corresponding entrywise bound. Taking the Frobenius norm over all matrix entries and applying Minkowski's inequality gives
\begin{equation}
\|\widehat{\bar A}-\bar A\|_F
\leq
\|\widehat A^{+}-A^{+}\|_F
+
\|\widehat A^{-}-A^{-}\|_F.
\label{eq:app-contrastive-matrix}
\end{equation}

This proves the lemma. Importantly, Equation~\eqref{eq:app-contrastive-matrix} does not require independence between $A^{+}$ and $A^{-}$, which are estimated from the same trajectory pool in Trace2Tower. The result therefore isolates a property of the proposed failure-suppression mechanism itself: the contrastive transformation does not amplify total affinity perturbation beyond the sum of its success- and failure-conditioned inputs.

% ============================================================
% Appendix B
% ============================================================

\subsection{Appendix B: Spectral Relaxation Characterization}
\label{app:spectral_characterization}

Consider a positive-degree connected component $\mathcal V_c$ with symmetric affinity $W^{(c)}$, degree matrix $D^{(c)}$, and normalized Laplacian
\begin{equation}
L^{(c)}
=
I-(D^{(c)})^{-1/2}
W^{(c)}
(D^{(c)})^{-1/2}.
\label{eq:app-normalized-laplacian}
\end{equation}

Because the component is connected, the smallest eigenvalue equals zero. Its normalized eigenvector is
\begin{equation}
q^{(c)}_1
=
\frac{(D^{(c)})^{1/2}\mathbf 1}
{\|(D^{(c)})^{1/2}\mathbf 1\|_2}.
\label{eq:app-trivial-eigenvector}
\end{equation}

For the eigengap-selected procedure count $r_c$, Trace2Tower uses the constrained relaxation
\begin{equation}
\begin{aligned}
\min_{Q\in\mathbb R^{m_c\times(r_c-1)}}
\quad
&\operatorname{tr}(Q^\top L^{(c)}Q) \\
\mathrm{s.t.}\quad
&Q^\top Q=I,
\qquad
Q^\top q^{(c)}_1=0.
\end{aligned}
\label{eq:app-spectral-relaxation}
\end{equation}

By the Rayleigh--Ritz/Courant--Fischer characterization \cite{horn2012matrix}, the minimizing subspace is spanned by the next $r_c-1$ eigenvectors. Hence one valid optimizer is
\begin{equation}
Q^{(c)}
=
[q^{(c)}_2,\ldots,q^{(c)}_{r_c}],
\label{eq:app-spectral-basis}
\end{equation}
and the minimum objective value is $\sum_{j=2}^{r_c}\lambda_j^{(c)}$.
Right multiplication by any orthogonal matrix gives an equivalent basis of the same subspace. Trace2Tower row-normalizes the retained representation before $K$-means, and orthogonal transformations preserve Euclidean distances. Therefore, the procedure partition is invariant to this global basis ambiguity. Zero-norm rows have no defined angular direction and are assigned to singleton groups.
The symmetrization $W=(\bar A+\bar A^\top)/2$ is used only for procedure discovery. Direction is retained in the original contrastive graph and reintroduced when compressed procedure sequences are used to build the directed Mid-level graph for strategy induction. Thus, the spectral step identifies stable procedure groups without discarding the execution order required by higher-level strategies.

% ============================================================
% Appendix C
% ============================================================

\section{Appendix C: Proof of EigenTrace Stability}
\label{app:eigentrace_stability}

Let $L=L^{(c)}$, $\widehat L=\widehat L^{(c)}$, $E=\widehat L-L$, $Q=[q_2,\ldots,q_{r_c}]$, and $\widehat Q=[\widehat q_2,\ldots,\widehat q_{r_c}]$. The theorem concerns perturbations on the same component node set; it therefore controls changes in graph weights and the corresponding invariant subspace rather than discrete changes in component membership.
Define the separation between the retained EigenTrace block and its complementary spectrum as
\begin{equation}
\Delta_c
=
\min
\left\{
\lambda_2-\lambda_1,\,
\lambda_{r_c+1}-\lambda_{r_c}
\right\}.
\label{eq:app-eigentrace-gap}
\end{equation}

Assume $\|E\|_2\leq\varepsilon_c$ and $\Delta_c>2\varepsilon_c$. A population-eigengap form of the Davis--Kahan sin-$\Theta$ theorem \cite{davis1970rotation,yu2015useful} gives
\begin{equation}
\|\sin\Theta(\widehat Q,Q)\|_2
\leq
\frac{2\varepsilon_c}{\Delta_c}.
\label{eq:app-eigentrace-operator}
\end{equation}

Since the retained subspace has dimension $r_c-1$,
\begin{equation}
\begin{aligned}
\|\sin\Theta(\widehat Q,Q)\|_F
&\leq
\sqrt{r_c-1}\,
\|\sin\Theta(\widehat Q,Q)\|_2 \\
&\leq
\frac{2\sqrt{r_c-1}\,\varepsilon_c}{\Delta_c}.
\end{aligned}
\label{eq:app-eigentrace-frobenius}
\end{equation}

Equation~\eqref{eq:app-eigentrace-frobenius} proves the theorem. The result is stated for the invariant subspace rather than individual eigenvectors because repeated or spaced eigenvalues inside the retained block can rotate its basis without changing the procedure-relevant representation. Together with Appendix~A, this establishes two distinct robustness properties: the contrastive transformation is stable with respect to errors in $A^{+}$ and $A^{-}$, and a sufficiently separated EigenTrace subspace is stable with respect to perturbations of the normalized graph.

% ============================================================
% Appendix D
% ============================================================

\section{Appendix D: Skill-Tower Induction and Deployment}
\label{app:induction_deployment}

\begin{table}[t]
\centering
\small
\setlength{\tabcolsep}{3.5pt}
\begin{tabular}{p{1.55cm}p{2.15cm}p{2.45cm}p{1.65cm}}
\toprule
Stage & Input & Main operation & Output \\
\midrule
Event abstraction & Raw trajectory & Segment and canonicalize & $\mathcal E$ \\
Graph induction & Canonical events & Semantic + transition + outcome evidence & $A^{+},A^{-}$ \\
Contrastive graph & $A^{+},A^{-}$ & Failure-aware transformation & $\bar A$ \\
Procedure induction & $\bar A$ & Component-wise spectral clustering & $\mathcal S^{(2)}$ \\
Strategy induction & Procedure sequences & Directed graph + SCC paths & $\mathcal S^{(3)}$ \\
Deployment & Skill tower & High plan + optional Mid support & Retrieved context \\
\bottomrule
\end{tabular}
\caption{Trace2Tower representations and their roles.}
\label{tab:app-pipeline}
\end{table}
Table~\ref{tab:app-pipeline} summarizes how each representation in the main paper is produced and what information it preserves.
Each environment interaction is serialized as $(g,o_t,a_t,o_{t+1},r_t)$ together with admissible-action information. Benchmark adapters map raw actions to domain event labels, and maximal consecutive runs with the same label form event segments. ALFWorld signatures retain normalized task context, neighboring event labels, segment length, and typed action templates; WebShop additionally retains compact page-state context. The reported representation dimensionality is 4,096.
A trajectory is labeled successful when its primary benchmark score is at least $0.999$. Node occurrence counts use each node at most once per trajectory, whereas transition counts retain every adjacent occurrence. For an observed transition $(u,v)$, semantic compatibility is
\begin{equation}
s_{uv}
=
\frac{1}{2}
+
\frac{h_u^\top h_v}
{2\|h_u\|_2\|h_v\|_2},
\label{eq:app-semantic-compatibility}
\end{equation}
with $s_{uv}=0$ if either representation has zero norm. Semantic similarity alone never creates an edge; the transition must have been observed.
The remaining graph statistics are
\begin{equation}
\begin{aligned}
\omega_{uv}
&=
\frac{c^+_{uv}+c^-_{uv}}
{1+c^+_{uv}+c^-_{uv}},
&
t_{uv}^{\pm}
&=
\frac{c_{uv}^{\pm}}
{\max(1,c_u^{\pm})},
\\
\rho_u^{\pm}
&=
\frac{n_u^{\pm}+1}
{n_u^{+}+n_u^{-}+2},
&
o_{uv}^{\pm}
&=
\sqrt{\rho_u^{\pm}\rho_v^{\pm}}.
\end{aligned}
\label{eq:app-graph-statistics}
\end{equation}

The affinities $A_{uv}^{\pm}$ and $\bar A_{uv}$ then follow Equations~\eqref{eq:outcome-conditioned-affinity} and \eqref{eq:contrastive-affinity}. No learned coefficient mixes semantic, transition, and outcome evidence.
For procedure induction, zero-degree nodes form singleton groups, while positive-degree nodes are processed component-wise. Components with fewer than three nodes remain single groups. Larger components use the eigengap-selected procedure count in Equation~\eqref{eq:eigengap-selection} and the EigenTrace representation in Equation~\eqref{eq:eigentrace-representation}, followed by $K$-means clustering. We use 20 initializations, a maximum of 300 iterations, and random seed 42.
Let $C(e)$ denote the procedure containing event segment $e$. Each trajectory is compressed into a procedure sequence
\begin{equation}
\pi_i
=
\left\langle
C(e_{i,j_1}),
C(e_{i,j_2}),
\ldots,
C(e_{i,j_{L_i}})
\right\rangle,
\label{eq:app-procedure-sequence}
\end{equation}
where $j_1,\ldots,j_{L_i}$ retain only the first event of each maximal run with the same procedure label. Thus, consecutive repetitions are removed while the original procedure order is preserved. Trace2Tower then rebuilds the transition, outcome, and contrastive graph over procedure sequences and collapses strongly connected components. Maximal paths observed in successful trajectories are retained only when every adjacent component pair has positive contrastive support, forming strategy-level skills. Structural membership is fixed before textual rendering, so the Skill Author cannot alter procedure membership, path order, or support statistics.

\begin{algorithm}[t]
\caption{Trace2Tower: Transition-Aware Skill-Tower Induction}
\label{alg:trace2tower}
\textbf{Input}: Trajectories $\mathcal D=\{(\tau_i,y_i)\}_{i=1}^{N}$\\
\textbf{Output}: tower $\mathcal T=\{\mathcal S^{(1)},\mathcal S^{(2)},\mathcal S^{(3)}\}$
\begin{algorithmic}[1]
\FOR{each $\tau_i\in\mathcal D$}
\STATE Construct canonical events $\mathcal E_i$ using Eq.~\eqref{eq:event-abstraction-overview}
\ENDFOR
\STATE Build event nodes $\mathcal V$ and success/failure statistics
\FOR{each observed transition $(u,v)$}
\STATE Compute $A_{uv}^{\pm}$ using Eq.~\eqref{eq:outcome-conditioned-affinity}
\STATE Compute $\bar A_{uv}$ using Eq.~\eqref{eq:contrastive-affinity}
\ENDFOR
\STATE Construct $W$, $D$, and $L$ using Eq.~\eqref{eq:normalized-eigentrace-graph}
\FOR{each component $\mathcal V_c$}
\STATE Select $r_c$ using Eq.~\eqref{eq:eigengap-selection}
\STATE Construct $Z^{(c)}$ using Eq.~\eqref{eq:eigentrace-representation}
\STATE Cluster $Z^{(c)}$ into procedure skills
\ENDFOR
\STATE Set canonical events as action skills $\mathcal S^{(1)}$
\STATE Collect induced procedures as $\mathcal S^{(2)}$

\STATE Retain maximal success-supported paths as $\mathcal S^{(3)}$
\STATE \textbf{return} $\mathcal T=\{\mathcal S^{(1)},\mathcal S^{(2)},\mathcal S^{(3)}\}$
\end{algorithmic}
\end{algorithm}

At deployment, both policies use the same induced Tower. The reported evaluator retrieves three High-level strategy references and rewrites them into a task-bound plan. \textsc{High-only} uses this plan alone. \textsc{Full} additionally retrieves procedure candidates for the rewritten plan steps, removes duplicates and candidates below cosine similarity $0.45$, and retains at most eight Mid cards after self-filtering.
The budgeted objective in Equation~\eqref{eq:hierarchical-retrieval} formalizes the desired hierarchical selection problem. The reported experiments instantiate this objective with the fixed two-stage policy above rather than solving the combinatorial argmax exactly. Mid retrieval is performed from the rewritten plan at episode start and is not refreshed after every environment transition.

% ============================================================
% Appendix E
% ============================================================

% ============================================================
% Appendix E
% ============================================================

\section{Appendix E: Experimental Configuration, Complexity, and Structural Analyses}
\label{app:experimental_analysis}

This appendix provides the computational complexity, experimental configuration, and additional analyses required to interpret the reported results. The implementation settings follow the method defined in the main paper and Appendix~\ref{app:induction_deployment}, while the additional analyses examine how the induced structure behaves under component removal, model transfer, and large-scale skill construction.

\subsection{E.1 Time and Memory Complexity}

Let $S$ be the number of raw interaction steps, $n$ the number of event segments, $m$ the number of quotient nodes, $e$ the number of observed quotient transitions, $d$ the embedding dimension, $m_c$ the size of component $c$, $R$ the number of procedure groups, and $P$ the total length of compressed procedure sequences. The graph remains sparse because semantic similarity is evaluated only for observed transitions. The dominant computational cost arises from the spectral stage. Since the reported configuration searches the full admissible eigengap of every nontrivial component, it computes a complete eigensystem for each component rather than using a truncated solver. Component-wise decomposition therefore changes the dense spectral cost from a monolithic $O(m^3)$ computation to $O(\sum_c m_c^3)$ and bounds peak dense memory by the largest component. The complexity in Table~\ref{tab:complexity} describes the reported configuration and does not assume an unreported partial eigensolver.

\begin{table}[t]
\centering
\small
\setlength{\tabcolsep}{3.5pt}
\begin{tabular}{lcc}
\toprule
Stage & Time & Working memory \\
\midrule
Event abstraction & $O(S)$ & $O(n)$ \\
Embedding / quotienting & $O(nd)$ & $O(nd)$ \\
Graph statistics & $O(n+ed)$ & $O(m+e)$ \\
Component eigensystems & $O(\sum_c m_c^3)$ & $O(\max_c m_c^2)$ \\
Procedure clustering & $O(\sum_c m_c r_c^2)$ & $O(\max_c m_c r_c)$ \\
Strategy graph / SCCs & $O(P+E_Rd+R+E_R)$ & $O(Rd+E_R)$ \\
\bottomrule
\end{tabular}
\caption{Dominant time and working-memory costs of the reported pipeline.}
\label{tab:complexity}
\end{table}

\subsection{E.2 Experimental Configuration}

\begin{table}[t]
\centering
\small
\setlength{\tabcolsep}{4pt}
\begin{tabular}{lrrrr}
\toprule
Benchmark & Train tasks & Rollouts/task & Trajectories & Test tasks \\
\midrule
ALFWorld & 310 & 4 & 1,240 & 134 \\
WebShop & 100 & 4 & 400 & 100 \\
\bottomrule
\end{tabular}
\caption{Construction and evaluation pools. Both benchmarks use a 20-step interaction horizon.}
\label{tab:app-benchmark-settings}
\end{table}

Table~\ref{tab:app-benchmark-settings} summarizes the benchmark construction and evaluation protocol.
ALFWorld evaluates all 134 solvable valid-unseen tasks, while WebShop uses a frozen 100-task test manifest. Construction and evaluation identifiers are disjoint in both environments. The same interaction horizon of 20 steps is used for all methods.
GPT-5.4 serves as the Skill Author and plan rewriter, while DeepSeek-V4-Flash is the default Skill User. The Skill User uses temperature 0 and a maximum of 512 output tokens per interaction turn. Skill rendering is schema constrained. The supplied experimental records confirm a 4,096-dimensional embedding representation processed in batches of 16. The exact embedding-model identifier is not recoverable from the supplied records and should therefore be taken from the final run metadata rather than inferred from representation dimensionality.

\begin{table}[t]
\centering
\small
\setlength{\tabcolsep}{4pt}
\begin{tabular}{lr}
\toprule
Setting & Value \\
\midrule
Outcome pseudo-counts & $1$ success + $1$ failure \\
Minimum nontrivial component & 3 nodes \\
Zero-degree tolerance & $10^{-12}$ \\
$K$-means initializations & 20 \\
$K$-means maximum iterations & 300 \\
Random seed & 42 \\
Retrieved High references & 3 \\
Mid candidates per plan step & 4 \\
Mid cosine threshold & $0.45$ \\
Maximum injected Mid cards & 8 \\
Per-transition Mid refresh & no \\
\bottomrule
\end{tabular}
\caption{Essential graph-induction and deployment settings.}
\label{tab:app-hyperparameters}
\end{table}

The essential method and deployment settings are summarized in Table~\ref{tab:app-hyperparameters}.
The graph-induction parameters correspond to the component-wise procedure induction described in Appendix~\ref{app:induction_deployment}. The deployment parameters specify the practical realization of the hierarchical retrieval objective in Equation~\eqref{eq:hierarchical-retrieval}. In particular, Full retrieves Mid-level support against the rewritten plan at episode initialization rather than refreshing the retrieval set after every environment transition.
For $R=3$ complete repetitions, each task metric is averaged within one repetition. Main tables report the arithmetic mean across repetitions and the sample standard deviation

\begin{equation}
s
=
\sqrt{
\frac{1}{R-1}
\sum_{r=1}^{R}
(x_r-\bar x)^2
}.
\label{eq:app-sample-standard-deviation}
\end{equation}

ALFWorld success uses the final environment completion indicator. WebShop exact success is the indicator that the official reward is at least $0.999$. Invalid actions include malformed, unavailable, or environment-rejected executable actions. Skill-context characters count only the skill information injected into the Skill User.
All automatic baselines use the same benchmark-specific trajectory pool, Skill User, evaluation manifest, environment implementation, and interaction horizon. ExpeL \cite{zhao2024expel}, SkillX \cite{wang2026skillx}, and Trace2Skill \cite{ni2026trace2skill} retain their native experience representations rather than being converted into the Trace2Tower interface.

\subsection{E.3 Structural Ablations}

\begin{table}[t]
\centering
\small
\setlength{\tabcolsep}{3.5pt}
\begin{tabular}{lccr}
\toprule
Variant & Retained signals & Mid/High & Success \\
\midrule
Full & S+T+O+C & 39/118 & $87.31\pm0.75$ \\
No Transition & S+O+C & 19/76 & 70.15 \\
No Outcome & S+T+C & 39/106 & 73.88 \\
No Contrastive & S+T+O & 10/44 & 73.88 \\
\bottomrule
\end{tabular}
\caption{ALFWorld structural ablations. S, T, O, and C denote semantic, transition, outcome, and contrastive evidence. Each ablation is one 134-task evaluation; Full is the three-run mean.}
\label{tab:structural_ablation_appendix}
\end{table}

The main paper reports overall benchmark performance, while Table~\ref{tab:structural_ablation_appendix} isolates the contribution of the graph signals that determine the induced hierarchy.
Removing transition evidence causes the largest performance degradation and reduces the number of induced procedure and strategy structures from $39/118$ to $19/76$. This result indicates that temporal dependencies provide information that cannot be recovered from semantic similarity and outcome evidence alone. Removing outcome evidence or the contrastive transformation also substantially lowers success. Together, these results support the use of success- and failure-conditioned behavioral topology rather than a purely semantic event graph.
Each structural ablation is evaluated once on the complete 134-task ALFWorld set, whereas Full reports the mean and sample standard deviation across three complete runs. The ablation values therefore diagnose the structural effect of removing individual signals and are not interpreted as run-to-run variance estimates.

\subsection{E.4 Cross-Model Transfer}

The cross-model experiment separates the model that constructs the Tower from the model that subsequently executes with it. With DeepSeek-V4-Flash as the Skill User, the GPT-5.4-authored Tower reaches 88.06\% success compared with 65.67\% for the DeepSeek-V4-Flash-authored Tower. Their paired outcomes contain 34 GPT-only wins, 4 Flash-only wins, and 96 ties, yielding an exact two-sided McNemar value of $p=6.04\times10^{-7}$.
With DeepSeek-V4-Pro as the Skill User, the corresponding success rates are 85.82\% and 79.10\%, with $p=0.150$. The improvement across both Skill Users indicates that the induced Tower is not tied to the model that authored its textual rendering, while the difference between the two Tower authors also shows that representation quality remains sensitive to the Skill Author.

\subsection{E.5 Hierarchy Scale and Construction Cost}

On ALFWorld, 1,240 construction trajectories produce 13,724 event segments and 3,764 reported quotient nodes. Trace2Tower subsequently induces 39 procedure skills and 118 strategy skills. Across 402 Full evaluation episodes, retrieval uses 53 distinct High skills and 30 distinct Mid skills, showing that deployment draws from a substantial fraction of both induced levels rather than repeatedly relying on a small fixed subset.
Skill construction uses 1,025,732 total GPT tokens, compared with 1,233,641 for SkillX. Trace2Tower therefore reduces total chat-generation tokens by 16.85\%. This reduction arises despite explicitly constructing procedure and strategy structure, indicating that structural induction does not require a larger language-model generation budget than iterative skill-library construction.
\begin{table}[t]
\centering
\small
\setlength{\tabcolsep}{4pt}
\begin{tabular}{p{1.3cm}p{2.55cm}p{3.7cm}}
\toprule
Edit & Target & Structural effect \\
\midrule
Split & Heterogeneous procedure & Creates evidence-specific children while retaining the parent \\
Merge & Compatible procedures & Contracts evidence and supported relations into one procedure \\
Promote & Supported Mid path & Adds a reusable strategy motif \\
Downweight & Harmful High skill & Adds a reversible retrieval penalty without deleting the skill \\
\bottomrule
\end{tabular}
\caption{Verifier-guided structural edit operators.}
\label{tab:app-edits}
\end{table}

% ============================================================
% Appendix F
% ============================================================

\section{Appendix F: Verifier-Guided Refinement and Feedback Evaluation}
\label{app:feedback_refinement}

This appendix describes how Trace2Tower updates an induced Tower using deployment feedback. Refinement operates on the frozen hierarchy without repeating event segmentation, graph construction, or spectral decomposition, allowing the learned structure to be corrected after deployment.

\subsection{F.1 Verifier-Guided Structural Refinement}

Candidate Towers are evaluated against the frozen Tower on the same task manifest using paired outcomes. Table~\ref{tab:app-edits} summarizes four edits: Split separates heterogeneous procedures, Merge combines compatible procedures, Promote elevates success-supported procedure paths, and Downweight reduces harmful strategy reuse without deleting the underlying structure.
Candidate edits are compared on semantic relevance, structural relevance, and verifier-feedback evidence. An edit is dominated if another is no worse on all three criteria and better on at least one. Only non-dominated candidates reach the deployment gate, avoiding an additional tuned scalar combination.
The held-out study uses 120 tasks per set and 10,000 paired bootstrap resamples. The accepted transaction splits one procedure, promotes eight strategy motifs, downweights one strategy, and performs no merge. Versioned updates enable direct comparison with the frozen Tower and rollback when the paired gate is not satisfied.

\subsection{F.2 Feedback Evaluation}

The frozen Tower achieves 80.42\% pooled success across the two held-out sets. Structural refinement increases this rate to 84.58\% with TF--IDF-based Pareto selection and 83.75\% with embedding-based selection.
The two relevance representations exhibit opposite set-level rankings: TF--IDF performs better on the first set, while embedding relevance performs better on the second. Their consistent improvement over the frozen Tower supports the contribution of editable structural relations and paired feedback rather than dependence on a single relevance representation.

\end{document}